\documentclass[journal,twoside,web]{ieeecolor}
\usepackage{generic}
\usepackage{cite}
\usepackage{amsmath,amssymb,amsfonts}
\usepackage{algorithmic}
\usepackage{graphicx}
\usepackage{algorithm,algorithmic}
\usepackage{hyperref}
\usepackage{booktabs}
\hypersetup{hidelinks=true}
\usepackage{textcomp}
\usepackage{multirow}
\usepackage{pifont}
\usepackage{xurl}

\def\BibTeX{{\rm B\kern-.05em{\sc i\kern-.025em b}\kern-.08em
    T\kern-.1667em\lower.7ex\hbox{E}\kern-.125emX}}
\begin{document}
\title{Fusion of Multiscale Features Via Centralized Sparse-attention Network for EEG Decoding}
\author{Xiangrui Cai, Shaocheng Ma, Lei Cao*, Jie Li*, Tianyu Liu*, and Yilin Dong \IEEEmembership{Member, IEEE}
\thanks{The work was supported by the Shanghai Science and Technology Project (No. 24YL1900200), National Natural Science Foundation of China (62472319), Light of Taihu Project of Wuxi Technology Commission (No. Y20242119), National Natural Innovation Research Group Project (82021002). (Xiangrui Cai and Shaocheng Ma contributed equally to this work.) (Corresponding author: Lei Cao, Jie Li, and Tianyu Liu, these authors contributed equally.)}
\thanks{Xiangrui Cai, Shaocheng Ma, Lei Cao, Tianyu Liu, and Yilin Dong are with the Department of Artificial Intelligence, Shanghai Maritime University, Shanghai 201306, China (e-mail: 202330310112@stu.shmtu.edu.cn; 202430310086@stu.shmtu.edu.cn; lcao@shmtu.edu.cn; liuty@shmtu.edu.cn; yldong@shmtu.edu.cn).}
\thanks{Jie Li is with the Translational Research Center, Shanghai Yangzhi Rehabilitation Hospital (Shanghai Sunshine Rehabilitation Center), School of Computer Science and Technology, Tongji University, Shanghai 200092, China (e-mail: nijanice@163.com.).
}}

\maketitle

\begin{abstract}
Electroencephalography (EEG) signal decoding is a key technology that translates brain activity into executable commands, laying the foundation for direct brain-machine interfacing and intelligent interaction. To address the inherent spatiotemporal heterogeneity of EEG signals, this paper proposes a multi-branch parallel architecture, where each temporal scale is equipped with an independent spatial feature extraction module. To further enhance multi-branch feature fusion, we propose a Fusion of Multiscale Features via Centralized Sparse-attention Network (EEG-CSANet), a centralized sparse-attention network. It employs a main-auxiliary branch architecture, where the main branch models core spatiotemporal patterns via multiscale self-attention, and the auxiliary branch facilitates efficient local interactions through sparse cross-attention. Experimental results show that EEG-CSANet achieves state-of-the-art (SOTA) performance across five public datasets (BCIC-IV-2A, BCIC-IV-2B, HGD, SEED, and SEED-VIG), with accuracies of 88.54\%, 91.09\%, 97.15\%, 96.03\%, and 90.56\%, respectively. Such performance demonstrates its strong adaptability and robustness across various EEG decoding tasks. Moreover, extensive ablation studies are conducted to enhance the interpretability of EEG-CSANet. In the future, we hope that EEG-CSANet could serve as a promising baseline model in the field of EEG signal decoding. The source code is publicly available at: \url{https://github.com/Xiangrui-Cai/EEG-CSANet}.
\end{abstract}


\section{Introduction}
\label{sec:introduction}
\IEEEPARstart{B}{rain–computer} interfaces (BCIs) enable direct communication between the human brain and external devices\cite{mcfarland2011brain} and have attracted increasing attention due to their broad application potential in human–computer interaction, motor rehabilitation, affective computing, fatigue detection, as well as disease diagnosis and treatment\cite{park2022brain}\cite{zhang2023adaptive}\cite{edelman2024non}. Electroencephalography (EEG), which measures voltage fluctuations generated by neural activity, has become the most widely used signal acquisition modality in contemporary BCI systems, owing to its noninvasive nature, high temporal resolution, and strong portability\cite{liu2025recent}. Currently, the dominant EEG paradigms include motor imagery (MI), event-related P300, and steady-state visual evoked potentials (SSVEP)\cite{yadav2025decoding}\cite{abiri2019comprehensive}. Owing to their distinct neural response characteristics, these paradigms have demonstrated strong effectiveness and adaptability across a wide range of application scenarios\cite{saibene2023eeg}\cite{qin2023application}.

However, efficiently and accurately decoding human intent remains a fundamental challenge in the BCI field, primarily due to the nonstationary nature of EEG signals, substantial inter-subject variability, and low signal noise ratio\cite{saibene2024deep}. To address this issue, a variety of machine learning (ML) and deep learning (DL) based approaches have been proposed to enhance feature extraction and classification performance of brain signals. Traditional ML-based methods typically consist of two sequential stages: feature extraction and classification. During the feature extraction stage, techniques such as Common Spatial Pattern (CSP), Differential Entropy (DE), and Power Spectral Density (PSD) are commonly employed. In the classification stage, classifiers including Linear Discriminant Analysis (LDA) and Support Vector Machine (SVM) have been shown to achieve competitive performance\cite{ang2012filter}\cite{zheng2015investigating}\cite{cao2022effective}. Nevertheless, these methods rely heavily on handcrafted feature engineering, which is labor-intensive and highly dependent on domain-specific prior knowledge, thereby limiting their robustness and generalization capability.

By comparison, deep learning based methods employ end-to-end models composed of multiple processing layers, enabling the automatic learning of intrinsic data representations and thereby substantially reducing the reliance on manual intervention and domain-specific preprocessing. With the rapid advancement of deep learning techniques, a wide range of deep learning models have been extensively applied to EEG decoding. Early studies introduced Deep ConvNet, Shallow ConvNet\cite{schirrmeister2017deep}, and EEGNet\cite{lawhern2018eegnet}, which significantly improved classification accuracy and model robustness, and have since become representative approaches for EEG feature extraction. Subsequently, Long Short-Term Memory (LSTM) networks\cite{zhang2019novel}, hybrid CNN–LSTM models\cite{wang20232d}, and multi-branch CNN architectures\cite{yang2021two} were proposed, further advancing EEG decoding performance. In recent years, driven by the success of Transformer architectures in sequence modeling, self-attention mechanisms have been introduced into EEG analysis. Models such as Conformer\cite{song2022eeg} and ATCNet\cite{altaheri2022physics} effectively integrate attention mechanisms with CNN architectures, leading to substantial improvements in feature representation capability and elevating EEG decoding accuracy to a new level. Consequently, models based on the “CNN + Transformer” paradigm have gradually become the dominant approach.

In recent years, multi-scale modeling approaches have gained widespread attention in EEG decoding tasks due to the complex spatiotemporal features of EEG signals\cite{zhou2025csbrain}\cite{wang2024cbramod}. Representative works, such as MSTFNet\cite{jin2024multiscale}, EEGTransNet\cite{ma2024attention}, MCMTNet\cite{yang2025mcmtnet}, and TMSA-Net\cite{zhao2025tmsa}, typically employ multi-scale temporal convolutions to extract features at different temporal resolutions, which are subsequently fused through concatenation or addition and then fed into subsequent spatial convolutional layers to capture inter-regional spatial patterns. However, these architectures implicitly rely on a critical assumption: that all temporal scales share the same spatial structure. This assumption contradicts neurophysiological evidence\cite{li2025frequency}\cite{tao2023adfcnn}, which indicates that different temporal scales correspond to distinct frequency components in EEG signals (e.g., $\delta$, $\theta$, $\alpha$, $\beta$, $\gamma$ bands), and that these frequency bands exhibit markedly different patterns of regional activation and functional connectivity in the brain\cite{pfurtscheller2006mu}\cite{pfurtscheller1997motor}. Given the aforementioned heterogeneous spatial correlations across different temporal scales, enforcing a shared spatial modeling mechanism across all scales may overlook frequency-specific spatial topologies, thus limiting the model’s representational capacity and decoding performance.

Based on this analysis, we argue that a multi-branch parallel architecture, in which each temporal scale is equipped with an independent spatial feature extraction module, can more precisely capture scale-specific channel cooperation patterns, avoid the confusion or loss of critical spatial information caused by shared spatial weights, and enhance both the flexibility and discriminative power of the extracted features.

At the same time, effectively integrating multi-branch features remains a key challenge in feature fusion\cite{zhao2025deep}\cite{han2025multi}. In this context, the cross-attention mechanism\cite{vaswani2017attention} has recently demonstrated strong potential for feature fusion due to its ability to model inter-branch dependencies\cite{cao2025bi}\cite{zhao2024feature}. Cross-attention dynamically captures semantic relationships between different branches through the interaction of queries, keys, and values, and performs information fusion across modalities or hierarchical levels based on attention weights. This mechanism not only strengthens the generalization capability of the learned representations but also facilitates extensive applications in fields such as vision–language processing and multimodal learning\cite{li2023blip}\cite{jian2024rethinking}\cite{shen2024icafusion}.

Motivated by these insights and the studies in\cite{chen2023learning}\cite{ma2025multiscale}, we propose an innovative fusion architecture that coordinates a primary branch with auxiliary branches. The primary branch employs a multi-scale multi-head self-attention mechanism (MSA) to extract and reinforce core spatiotemporal features, while multiple auxiliary branches utilize a multi-scale sparse multi-head cross-attention mechanism (MSCA) to focus on local details at specific scales. During the fusion stage, MSCA guides the auxiliary branches to interact only with semantically relevant key local regions in the primary branch, effectively avoiding computational redundancy and noise interference caused by global associations. Specifically, multi-scale feature extraction integrates contextual information at different granularities, enriching the overall representation; the cross-attention mechanism ensures precise semantic alignment between primary and auxiliary branches, maintaining consistency between local details and global structure; and the sparsity mechanism further constrains the interaction scope, suppressing irrelevant regions and reducing computational overhead. The synergistic effect of these components enables MSCA to efficiently refine key local features under the semantic guidance of the primary branch, complementing MSA and achieving effective and precise feature aggregation, thereby substantially enhancing the model’s representational capacity and robustness. 

The contributions of this paper are as follows.

1. We propose a novel Fusion of Multiscale Features via Centralized Sparse-attention Network (EEG-CSANet), which effectively addresses the loss of channel discriminative information caused by coarse-grained feature fusion in the traditional multiscale temporal feature integration process.

2. We propose a feature fusion architecture that synergistically integrates a primary branch with an auxiliary branch. The primary branch employs a multi-scale multi-head self-attention mechanism to enhance the modeling of core spatiotemporal patterns, while the auxiliary branch leverages a multi-scale sparse multi-head cross-attention mechanism to enable efficient and precise feature interactions with the local key regions of the primary branch.

3. We conduct experiments on five public datasets (BCIC-IV-2a, BCIC-IV-2b, HGD, SEED, and SEED-VIG), and the results demonstrate that our method consistently achieves superior classification accuracy and generalization performance compared to existing approaches, thereby validating its effectiveness and robustness.

4. To advance further research in EEG decoding, we have open-sourced our proposed model on GitHub: \url{https://github.com/Xiangrui-Cai/EEG-CSANet}.

\section{Methods}
\label{sec:methods}

\begin{figure*}[!t]
\centering
\includegraphics[width=7in]{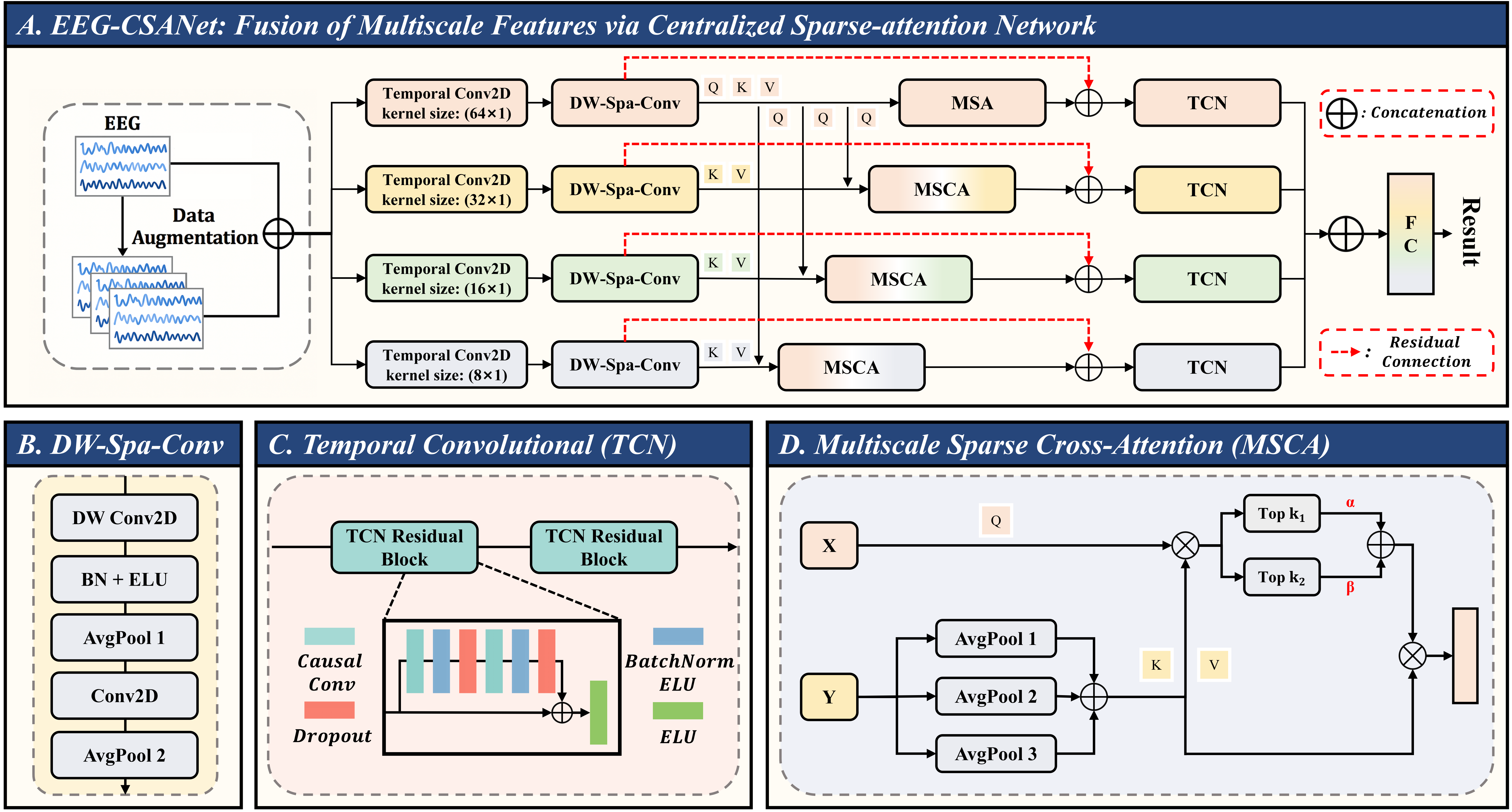}
\caption{The overall architecture of EEG-CSANet, including data augmentation, multi-branch convolution, feature fusion architecture, and temporal convolutional.}
\label{fig_1}
\end{figure*}

\begin{table}[htbp]
\centering
\caption{Hyperparameters of Different Modules.}
\label{tab_1}
\small
\resizebox{\linewidth}{!}{
\renewcommand{\arraystretch}{1.05}
\begin{tabular}{*{3}{c}}
\toprule
\textbf{Module} & \textbf{Hyperparameters} & \textbf{Settings} \\
\midrule
Data Augmentation & Segment ($S$) & 8 \\
\midrule
\multirow{2}{*}{Temporal Conv2D} 
& Kernel Size ($K_1,K_2,K_3,K_4$) & (64, 32, 16, 8) \\
& Filters ($F_1,F_2,F_3,F_4$) & (16, 16, 16, 16) \\
\midrule
\multirow{6}{*}[-0.8ex]{DW-Spa-Conv}
& DW Kernel Size ($K_5$) & (C, 1) \\
& DW Filters ($F_5$) & 16 \\
& Depth multiplier ($D$) & 2 \\
& Pooling Size ($P_1,P_2$) & (8, 7) \\
& Spa Filters ($F_6$) & 32 \\
& Spa Filters ($K_6$) & 32 \\
& Dropout & 0.5 \\
\midrule
\multirow{4}{*}{MSCA} 
& Pooling Size ($Q_1,Q_2,Q_3$) & (3, 5, 7) \\
& Pooling Padding ($G_1,G_2,G_3$) & (1, 2, 3) \\
& Number of Heads ($h$) & 8 \\
& Top-k ($k_1,k_2$) & (2, 3) \\
\midrule
\multirow{3}{*}[-1ex]{TCN}
& Dilation factor ($d_1,d_2$) & (1, 2) \\
& Kernel size ($K_t$) & 4 \\
& Filters ($F_t$) & 32 \\
& Dropout & 0.3 \\
\bottomrule
\end{tabular}}
\end{table}

The proposed EEG-CSANet architecture is illustrated in Figure~\ref{fig_1}. Initially, a same category signal segmentation and reconstruction (S\&R)\cite{lotte2015signal} method is employed for data augmentation of EEG signals. The original EEG data are then combined with the augmented data to expand the training sample size. Subsequently, the signals are processed by a multiscale temporal feature extraction module. To avoid losing inter-channel interaction information at different scales, a multi-branch structure is adopted, which separately extracts deep channel features for each time scale. To effectively integrate EEG signals extracted from multiple branches, we have innovatively designed a fusion mechanism: the main branch employs a multiscale multi-head self-attention mechanism, while auxiliary branches introduce a multiscale sparse multi-head cross-attention mechanism, achieving efficient information exchange between branches. The integrated features are then fed into a temporal convolutional network to further extract higher-level temporal dynamic features. To prevent the degradation of temporal information within deep networks, residual structures are incorporated into the convolutional modules. Finally, a fully connected layer is used to complete the classification task, the specific parameters are shown in Table~\ref{tab_1}. The following sections will provide a detailed description of each component of the EEG-CSANet.

\subsection{Data augmentation}
We use the method of S\&R to perform data augmentation on EEG signals, which is a commonly used data augmentation method in MI\cite{song2022eeg}\cite{ma2024attention}\cite{yang2025mcmtnet}. Specifically, we let $ X_i \in \mathbb{R}^{B \times C \times T} $ denote the original EEG signal of the $ i $-th class, where $ i \in \{1, \dots, L\} $ , $ L $ is the total number of classes, and $B$, $C$, and $T$ denote the batch size, number of channels, and number of time samples, respectively. The signal $ X_i $ is uniformly divided into $ S $ segments along the time dimension. New samples $ X_i' $ are then generated by randomly selecting and reorganizing these segments while preserving their original temporal order. This augmentation is applied separately within each training batch, generating an amount of synthetic data equal in size to the original input $ X_i $. Therefore, the model input for each batch after augmentation is given by:
\begin{align}
X_{\mathrm{Input}} = \mathrm{Concat} \left(X , X' \right), X = \sum_{k=1}^{L} X_i ,
\end{align}
where $X$ represents the original input signals in the batch, $ X_{\mathrm{Input}} $ denotes the complete input fed into the subsequent model, and $ {\mathrm{Concat(\cdot)}} $ denotes concatenation along the first dimension of the data.

\begin{figure*}[!t]
\centering
\includegraphics[width=6.5in]{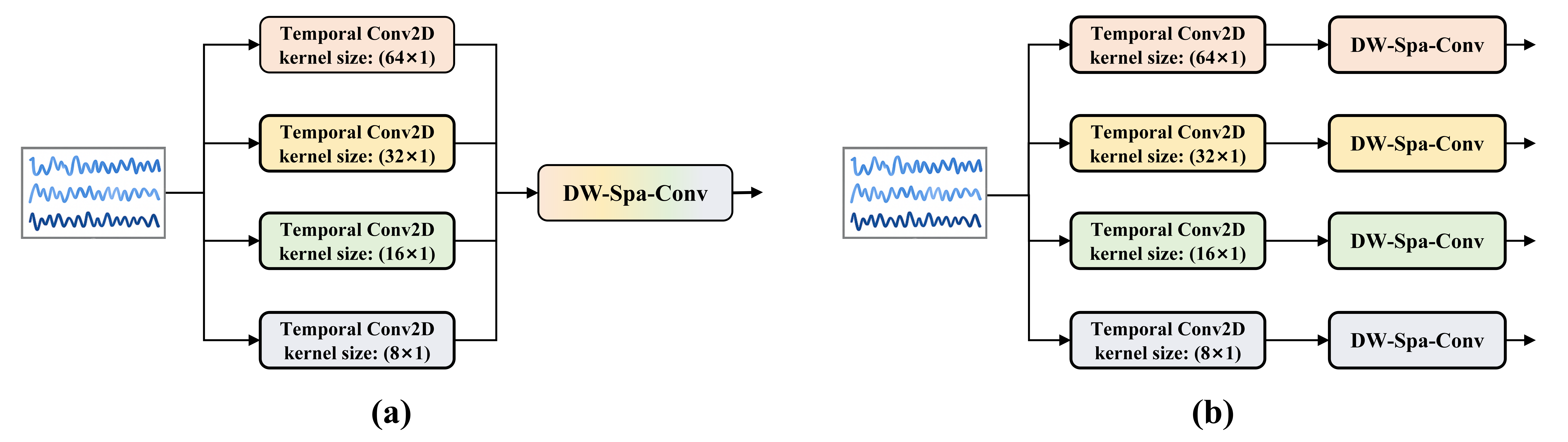}
\caption{Traditional multiscale feature extraction method and the proposed multibranch feature extraction method. (a) Traditional multiscale feature extraction method. (b) Proposed multibranch feature extraction method.}
\label{fig_2}
\end{figure*}

\subsection{Multi-branch convolution}
Previous studies\cite{jin2024multiscale}\cite{ma2024attention}\cite{yang2025mcmtnet}\cite{zhao2025tmsa} have tended to employ multiscale temporal convolution fusion methods to extract features from EEG signals, as illustrated in Figure~\ref{fig_2}. This approach concatenates multiscale temporal features to jointly extract channel-wise features. However, we contend that directly concatenating multiscale temporal features for channel-wise feature extraction may not effectively capture these distinct spatial patterns. Therefore, we propose a multi-branch architecture to separately process and fuse features from different temporal scales, as shown in Figure~\ref{fig_2}. In this architecture, each branch is responsible for handling features at a specific temporal scale or frequency band, thereby enabling more accurate capture and representation of the distinct spatial characteristics present across different frequency bands.

The combined structure of temporal convolution and depthwise separable spatial convolution (DW-Spa-Conv) resembles that of EEGNet\cite{lawhern2018eegnet}. Specifically, the input signal $ X_{\mathrm{Input}} $ first passes through a temporal convolutional module with kernel size $(K_i, 1)$, where $i \in \{1,2,3,4\}$ corresponds to the four branches, to extract temporal features for each channel. This is followed by a depthwise separable convolution with kernel size $(C, 1)$, where the depth multiplier $D$ controls the number of output feature maps per input channel, enabling spatial feature extraction. Next, the features are downsampled via the first average pooling layer with kernel size $(P_1, 1)$. The pooled features are then fed into a subsequent convolutional module with kernel size $(K_5, 1)$; we regard this module as facilitating further channel-wise interaction, thus enhancing the extraction of high-dimensional spatio-temporal features. Finally, a second average pooling layer with kernel size $(P_2, 1)$ performs additional downsampling. As a result, the output of each of the four convolutional branches is $Z_i \in \mathbb{R}^{B \times U_i \times T_0}$:
\begin{align}
U_i &= F_i \times D, \\
T_0 &= \frac{T}{P_1 \times P_2},
\end{align}
where $F_i$ denotes the number of convolutional kernels in the $i$-th temporal branch, and $T_0$ is the number of time steps after downsampling.

\begin{table*}[htbp]
\centering
\caption{Details of Different Datasets.}
\label{tab_2}
\renewcommand{\arraystretch}{1.15}
\setlength{\tabcolsep}{15pt} 
\small
\begin{tabular}{cccccc}
\toprule
\textbf{Dataset} & \textbf{Participants} & \textbf{Paradigm} & \textbf{Sampling rate (Hz)} & \textbf{Channels} & \textbf{Classes} \\
\midrule
BCIC-IV-2A & 9  & MI      & 250 & 22 & 4 \\
BCIC-IV-2B & 9  & MI      & 250 & 3  & 2 \\
HGD        & 14 & MI      & 250 & 44 & 4 \\
SEED       & 15 & Emotion & 200 & 64 & 3 \\
SEED-VIG   & 23 & Fatigue & 200 & 17 & 2 \\
\bottomrule
\end{tabular}
\end{table*}

\subsection{Feature fusion architecture}
To effectively fuse the features $Z_i$ obtained from the four branches, we propose a feature fusion architecture with collaboration between a main-branch multiscale multi-head self-attention mechanism and multiscale sparse multi-head cross-attention mechanisms for the auxiliary branches. 

Specifically, we designate $Z_1$ as the main branch and $Z_2$, $Z_3$, $Z_4$ as auxiliary branches. For the main branch $Z_1$, we adopt a relatively large convolutional kernel $K_1$ for temporal feature extraction, aiming to capture extensive global spatio-temporal dependencies. In contrast, the auxiliary branches $Z_2$, $Z_3$, and $Z_4$ use smaller kernels $K_2$, $K_3$, $K_4$ to preserve finer local spatio-temporal patterns. Due to the constrained receptive fields of small kernels, they are often inadequate for modeling long-range dependencies and can overlook crucial global structural information. To address this issue, we introduce a multiscale sparse multi-head cross-attention mechanism\cite{chen2023learning} that enhances global perception while capturing local fine-grained details, enabling collaborative fusion between the main and auxiliary branches.

As illustrated in Figure~\ref{fig_1}.D, $X$ and $Y$ are the inputs to the Multiscale Sparse Cross-Attention (MSCA) module. To capture multiscale information from the data, we first apply three average pooling operations with different kernel sizes, and then sum the resulting features to obtain $Y'$:
\begin{align}
Y' &= \sum_{i=1}^{3} Q_i(Y), \quad i \in \{1, 2, 3\},
\end{align}
where $Q_i$ denotes the $i$-th average pooling operation, and $Y'$ denotes the output of the pooling layer. 
Note that $Y$, $Y_i$, and $Y'$ all have the same spatial dimensions.

We then project the input $X$ into the Query matrix and the transformed input $Y'$ into the Key and Value matrices as follows:
\begin{equation}
    Q = X W_q, \quad
    K = Y' W_k, \quad
    V = Y' W_v,
\end{equation}
where $W_q$, $W_k$, and $W_v$ are learnable weight matrices. 

The attention score matrix $A$ is subsequently computed using scaled dot-product attention\cite{vaswani2017attention}:
\begin{equation}
    A = \frac{Q K^\top}{\sqrt{d_k}},
\end{equation}
where $d_k$ denotes the dimensionality of the key vectors.

Before applying the softmax function, we introduce a $\mathrm{Top}\text{-}k$ sparsification operation\cite{chen2023learning}, which discards the smallest $1-k$ proportion of values in each row of $A$. This effectively mitigates the influence of noise from auxiliary branches on feature extraction, as illustrated in Figure~\ref{fig_1}$.D$. For the softmax operation, these discarded values are replaced with $-\infty$, causing them to be transformed into zero after softmax, similar to a selective masking mechanism. The operation is expressed as:
\begin{align}
A' &= \mathrm{softmax}(\mathrm{Top}\text{-}k(A)),
\end{align}

We apply the $\mathrm{Top}\text{-}k$ operation with two different ratios, controlled by parameters $k_1$ and $k_2$, and introduce two learnable scalars $\alpha$ and $\beta$ to adaptively balance their contributions, it is a strategy inspired by\cite{ma2025multiscale}. Thus, the final attention output is computed as:
\begin{align}
\mathrm{Attention} = & \, \alpha \cdot \mathrm{softmax}(\mathrm{Top}\text{-}k_1(A)) \cdot V \nonumber \\
& + \beta \cdot \mathrm{softmax}(\mathrm{Top}\text{-}k_2(A)) \cdot V,
\end{align}

The output of the multi-head attention module is then concatenated across heads:
\begin{align}
\mathrm{MHA} &= \mathrm{Concat}\left( \mathrm{Attention}_0, \mathrm{Attention}_1, \dots, \mathrm{Attention}_{h-1} \right),
\end{align}
where $h$ denotes the number of attention heads.

This design enables the effective fusion of information from the main branch $X$ and auxiliary branches $Y$, thereby achieving cross-branch feature enhancement. Notably, sparsification is omitted in the main branch, which instead employs a multiscale multi-head self-attention mechanism to retain global temporal patterns. The combination of multiscale pooling and sparsification, however, may result in a significant loss of temporal information. To address this, we adopt a residual connection\cite{he2016deep} to retain the original features:
\begin{align}
M_i &= Z_i + \mathrm{MHA}_i,
\end{align}
where $M_i$ is the output of the $i$-th branch, $Z_i$ is the output from the DW-Spa-Conv block, and $\mathrm{MHA}_i$ is the corresponding MHA output.

\subsection{Temporal Convolutional}
The TCN structure we use follows the design of\cite{altaheri2022physics}, with its core architecture shown in Figure~\ref{fig_1}.C. This module adopts a double residual structure, aiming to deeply extract high-level temporal features from brain electrical signals. The dilation factors for the two residual blocks are $d_1$ and $d_2$, respectively. Batch normalization and ELU activation functions are then applied to enhance training stability. Finally, after concatenating the data, it is fed into a fully connected layer for classification, yielding the final classification result:
\begin{equation}
\hat{y} = \mathrm{Linear}\left(\mathrm{Concat} (P_1, P_2, P_3, P_4)\right),
\end{equation}
where $\mathrm{Linear}$ represents the final linear classification layer, and $ P_i$ represents the output features of the $i$-th layer of the TCN.

\section{Experiment Setup}
This section presents the five datasets and performance metrics adopted in the EEG-CSANet framework. As shown in the Table~\ref{tab_2}, these EEG datasets were collected using different acquisition devices, experimental paradigms, numbers of subjects, and sample sizes, thereby enabling a fair validation of the generalization ability of our method.

\subsection{BCIC-IV-2A}
BCIC-IV-2A\cite{brunner2008bci} provided by the Graz University of Technology, consists of recordings from nine healthy subjects. Each subject performed four MI tasks, including left hand, right hand, foot, and tongue movements. EEG signals were recorded using 22 Ag/AgCl electrodes at a sampling rate of 250 Hz. For each subject, two sessions were recorded on different days, serving as the training and testing sets, respectively. Each session comprises 288 trials, i.e., 72 trials per task. In our study, a 4-second time window was adopted for each trial\cite{song2022eeg}.

\subsection{BCIC-IV-2B}
BCIC-IV-2B\cite{leeb2008bci} also includes recordings from nine healthy subjects. Each subject performed two MI tasks, namely left-hand and right-hand movements. EEG signals were recorded using three bipolar electrodes (C3, Cz, and C4) at a sampling rate of 250 Hz. A 4-second time window was adopted for each trial\cite{song2022eeg}. Each subject completed five sessions: the first two sessions without feedback, each containing 120 trials, and the subsequent three sessions with feedback, each containing 160 trials. In our experiments, the first three sessions were used for training and the last two sessions for testing, as done in\cite{ma2024attention}.

\subsection{HGD}
HGD\cite{schirrmeister2017deep} consists of recordings from 14 healthy subjects. Each subject performed four MI tasks, including left-hand, right-hand, both-feet movements, and rest (where a visual cue identical to that of the other tasks was presented on the screen, but subjects were instructed not to perform any movement). The original EEG signals were recorded using 128 channels at a sampling rate of 500 Hz, with a 4-second time window for each trial. Following\cite{zhao2025tmsa}, we selected 44 channels covering the motor cortex for EEG decoding, downsampled the signals to 250 Hz, and applied standard normalization. For each subject, the dataset includes both training and testing sessions: the training set comprises approximately 880 trials per subject, while the testing set contains about 160 trials per subject.

\subsection{SEED}
SEED\cite{zheng2018emotionmeter} comprises emotion-related EEG signals from 15 subjects. In the experimental paradigm, 15 film clips were employed to induce three distinct emotional states, namely positive, neutral, and negative. For each subject, data acquisition was conducted across three sessions, with an interval of approximately one week between consecutive sessions. EEG signals were recorded from 62 electrodes at a sampling rate of 1000 Hz and subsequently downsampled to 200 Hz. The continuous signals were segmented into non-overlapping 1-second windows, resulting in 3394 trials per session. All EEG recordings were further processed using a band-pass filter in the range of 4-47 Hz. Consistent with the experimental protocol described in\cite{song2022eeg}, five-fold cross-validation was performed within a session.

\subsection{SEED-VIG}
SEED-VIG\cite{zheng2017multimodal} comprises driving fatigue related EEG signals from 23 subjects. Participants performed simulated driving tasks in a virtual-reality environment covering a variety of weather and road conditions, with a total recording duration of 118 minutes. EEG signals were acquired from 17 channels at an initial sampling rate of 1000 Hz and subsequently downsampled to 200 Hz. To suppress electronic interference, the recordings were band-pass filtered in the 1-50 Hz range and then normalized.

During the experiment, the PERCLOS index was computed every 8 seconds, and the data were segmented into trials of the same duration, yielding 885 trials per experimental run. PERCLOS is a quantitative metric used to assess the level of driver fatigue, which is defined as:
\begin{equation}
    \mathrm{PERCLOS} = \frac{\mathrm{blink} + \mathrm{close}}{\mathrm{interval}}
\end{equation}
where ${\mathrm{blink}}$, ${\mathrm{close}}$, and ${\mathrm{interval}}$ denote the blink duration, eye-closure duration, and total experimental time, respectively. 

Following\cite{zheng2017multimodal}, a threshold of 0.35 was used to distinguish between fatigued and alert states. Five-fold cross-validation was employed to partition the data into train and test sets.

\subsection{Experimental Details}
All datasets are implemented using Python 3.10 and the PyTorch framework, and experiments are conducted on an NVIDIA RTX 2080Ti GPU. The cross-entropy loss function and the Adam optimizer are used with a learning rate set to 0.0009. To ensure fair comparison with baseline models, the number of training epochs for BCIC-IV-2A, BCIC-IV-2B, and HGD is set to 2000, with a batch size of 64. For SEED and SEED-VIG, the batch size is set to 128\cite{chen2025driver}, with training epochs of 1000 and 500, respectively. The parameter $S$ in S\&R is set to 8. Following\cite{song2022eeg}, no data augmentation is applied to SEED. Since each trial in SEED is segmented into 1s windows, the model's pooling sizes ($P_1,P_2$) are set to (4, 4), while other datasets follow the specifications in Table~\ref{tab_2}. To ensure reproducibility, a fixed random seed is used throughout all experiments.

To address random variability and class imbalance, accuracy, standard deviation (STD), and the Kappa coefficient are employed as comprehensive metrics for evaluating the model's classification performance.

Assuming a classification task with $n$ categories, the accuracy (ACC) was defined as:
\begin{equation}
    \mathrm{ACC} = \frac{\sum_{i=1}^{n} \mathrm{TP}_i}{\sum_{i=1}^{n} (\mathrm{TP}_i + \mathrm{FP}_i)}
\end{equation}
where ${\mathrm{TP_i}}$ and ${\mathrm{FP_i}}$ denote the number of true positives and false positives for class $i$, respectively.

Standard deviation (STD) can be defined as:
\begin{equation}
\mathrm{STD} = \sqrt{\frac{1}{\mathrm{M}} \sum_{j=1}^{\mathrm{M}} \left( \mathrm{ACC}_j - \overline{\mathrm{ACC}} \right)^2},
\end{equation}
where \( \mathrm{M} \) denotes the number of independent runs, \( \mathrm{ACC}_j \) is the accuracy obtained in the \( j \)-th subject, and \( \overline{\mathrm{ACC}} \) represents the mean accuracy across all subjects.

A lower STD indicates more consistent and robust classification performance, while a higher STD suggests greater sensitivity to data partitioning or random initialization.

Kappa coefficient is defined as:
\begin{equation}
    \kappa = \frac{p_o - p_e}{1 - p_e}
\end{equation}
where \( p_o \) represents the observed agreement (i.e., the proportion of correctly predicted samples), and \( p_e \) denotes the expected agreement by chance.

\section{Result}
To ensure scientific rigor and a fair comparison, the proposed method was evaluated against several state-of-the-art (SOTA) approaches.

For the three MI-based datasets (BCIC-IV-2A, BCIC-IV-2B, and HGD), we compared our method with the baseline model EEGNet\cite{lawhern2018eegnet} as well as several SOTA models proposed between 2023 and 2025, including EEG-Conformer\cite{song2022eeg}, ATCNet\cite{altaheri2022physics}, ADFCNN\cite{tao2023adfcnn}, EEG TransNet\cite{ma2024attention}, MSTFNet\cite{jin2024multiscale}, EISATC-Fusion\cite{liang2024eisatc}, MCMTNet\cite{yang2025mcmtnet}, and TMSA-Net\cite{zhao2025tmsa}.

\begin{table*}[htbp]
\centering
\caption{Performance Comparison on BCIC-IV-2A. ($*$: p < $0.05$, $**$: p < $0.01$, $*\!*\!*$: p < $0.001$)}
\label{tab_3}
\small
\resizebox{\textwidth}{!}{
\begin{tabular}{cccccccccccccc}
\toprule
\textbf{Methods} & \textbf{Year} & \textbf{A1} & \textbf{A2} & \textbf{A3} & \textbf{A4} & \textbf{A5} & \textbf{A6} & \textbf{A7} & \textbf{A8} & \textbf{A9} & \textbf{ACC} & \textbf{STD} & \textbf{Kappa} \\
\midrule
EEGNet\textsuperscript{***} & 2018 & 82.91 & 66.49 & 87.29 & 59.39 & 64.88 & 60.59 & 72.81 & 81.06 & 85.57 & 73.44 & 11.02 & 0.6423 \\
EEG-Conformer\textsuperscript{**} & 2023 & 88.19 & 61.46 & 93.40 & 78.13 & 52.08 & 65.28 & 92.36 & 88.19 & 88.89 & 78.66 & 15.30 & 0.7155 \\
ATCNet\textsuperscript{***} & 2023 & 85.07 & 68.75 & 96.53 & 83.68 & 77.78 & 72.22 & 86.81 & 86.46 & 89.31 & 82.96 &  8.66 & 0.7840 \\
ADFCNN\textsuperscript{**} & 2024 & 89.42 & 71.12 & 95.61 & 82.43 & 73.42 & 71.88 & 90.97 & 87.50 & 86.81 & 83.24 &  9.05 & 0.7733 \\
EISATC-Fusion\textsuperscript{**} & 2024 & 85.07 & 73.26 & 95.49 & 87.15 & 81.94 & 73.96 & 93.06 & 85.76 & 85.42 & 84.57 &  \textbf{7.48} & 0.7942 \\
EEG-TransNet\textsuperscript{**} & 2024 & 88.89 & 64.93 & 96.18 & 85.42 & 82.64 & 73.61 & \textbf{95.14} & 90.28 & 88.19 & 85.03 & 10.12 & 0.8004 \\
MSTFNet\textsuperscript{**} & 2024 & 90.63 & 69.89 & 97.22 & 79.56 & 79.86 & 68.40 & 90.97 & 86.11 & 89.93 & 83.62 &  9.91 & 0.7900 \\
MCMTNet\textsuperscript{*} & 2024  & 89.70 & \textbf{73.43} & 95.01 & 82.38 & 80.79 & 70.88 & 92.29 & 87.94 & 88.01 & 84.49 &  8.28 & 0.7930 \\
TMSA-Net\textsuperscript{**} & 2025  & 87.50 & 64.24 & 96.18 & 84.03 & 79.86 & 67.71 & 93.06 & 90.79 & 85.42 & 83.20 & 10.94 & 0.7762 \\
\midrule
\textbf{EEG-CSANet} & \textbf{2025} & \textbf{94.44} & 71.88 & \textbf{98.26} & \textbf{91.32} & \textbf{84.03} & \textbf{79.51} & 93.75 & \textbf{92.36} & \textbf{91.32} & \textbf{88.54} & 8.41 & \textbf{0.8472} \\
\bottomrule
\end{tabular}}
\end{table*}

\begin{table*}[htbp]
\centering
\caption{Performance Comparison on BCIC-IV-2B. ($*$: p < $0.05$, $**$: p < $0.01$, $*\!*\!*$: p < $0.001$)}
\label{tab_4}
\small
\resizebox{\textwidth}{!}{
\begin{tabular}{cccccccccccccc}
\toprule
\textbf{Methods} & \textbf{Year} & \textbf{B1} & \textbf{B2} & \textbf{B3} & \textbf{B4} & \textbf{B5} & \textbf{B6} & \textbf{B7} & \textbf{B8} & \textbf{B9} & \textbf{ACC} & \textbf{STD} & \textbf{Kappa} \\
\midrule
EEGNet\textsuperscript{**} & 2018  & 75.94 & 57.64 & 58.43 & 98.13 & 81.25 & 88.75 & 84.06 & 93.44 & 89.69 & 80.81 & 14.46 & 0.6096 \\
EEG-Conformer\textsuperscript{*} & 2023 & 82.50 & 65.71 & 63.75 & 98.44 & 86.56 & 90.31 & 87.81 & 94.38 & 92.19 & 84.63 & 12.18 & 0.6926 \\
ATCNet\textsuperscript{*} & 2023 & 79.38 & 75.00 & \textbf{88.75} & 98.13 & 96.56 & 88.13 & 94.38 & 94.69 & 85.31 & 88.93 &  \textbf{7.94} & 0.7785 \\
ADFCNN\textsuperscript{**} & 2024 & 78.13 & 72.14 & 87.50 & 98.44 & 97.81 & 90.00 & 92.81 & 92.81 & 90.63 & 88.92 &  8.69 & 0.7749 \\
EISATC-Fusion\textsuperscript{**} & 2024 & 75.00 & 72.86 & 86.56 & 96.88 & 97.81 & 84.38 & 94.06 & 93.75 & 86.88 & 87.58 &  9.07 & 0.7515 \\
EEG-TransNet\textsuperscript{**} & 2024 & 79.06 & 70.71 & 87.81 & 98.44 & 96.88 & \textbf{91.56} & 91.88 & 95.63 & 90.00 & 89.11 &  8.98 & 0.7822 \\
MSTFNet\textsuperscript{*} & 2024 & 82.72 & 73.26 & 82.34 & 98.75 & 97.44 & 90.86 & 92.51 & 94.52 & 90.99 & 89.27 &  8.27 & 0.7800 \\
TMSA-Net\textsuperscript{*} & 2025 & 82.81 & 71.07 & 87.81 & 98.13 & 98.44 & \textbf{91.56} & 94.38 & 95.63 & 87.50 & 89.70 &  8.74 & 0.7940 \\
\midrule
\textbf{EEG-CSANet} & \textbf{2025} & \textbf{83.13} & \textbf{73.57} & 88.13 & \textbf{99.38} & \textbf{99.69} & 91.25 & \textbf{95.31} & \textbf{96.88} & \textbf{92.50} & \textbf{91.09} & 8.48 & \textbf{0.8218} \\
\bottomrule
\end{tabular}}
\end{table*}

\begin{table}[htbp]
\centering
\caption{Performance Comparison on HGD. \\($*$: p < $0.05$, $**$: p < $0.01$, $*\!*\!*$: p < $0.001$)}
\label{tab_5}
\small
\renewcommand{\arraystretch}{1.15}
\setlength{\tabcolsep}{9.3pt} 
\begin{tabular}{cccccc}
\toprule
\textbf{Methods} & \textbf{Year} & \textbf{ACC} & \textbf{STD} & \textbf{Kappa} \\
\midrule
EEGNet\textsuperscript{***} & 2018 & 88.87 & 6.53 & 0.8531 \\
EEG-Conformer\textsuperscript{***} & 2023 & 91.92 & 5.90 & 0.8824 \\
ATCNet & 2023 & 95.54 & 3.54 & 0.9405 \\
ADFCNN & 2024 & 95.17 & 7.74 & 0.9381 \\
EEG-TransNet\textsuperscript{*} & 2024 & 94.82 & 4.17 & 0.9309 \\
MCMTNet & 2025 & 95.73 & 3.23 & 0.9430 \\
TMSA-Net & 2025 & 95.90 & 8.52 & 0.9452 \\
\midrule
\textbf{EEG-CSANet} & \textbf{2025} & \textbf{97.15} & \textbf{2.97} & \textbf{0.9627} \\
\bottomrule
\end{tabular}
\end{table}

\begin{table*}[htbp]
\centering
\caption{Performance comparison on SEED and SEED-VIG.}
\label{tab_6}
\small
\renewcommand{\arraystretch}{1.15}
\setlength{\tabcolsep}{16.4pt} 
\begin{tabular}{cccc|cccc}
\toprule
\multicolumn{4}{c|}{\textbf{SEED}} & \multicolumn{4}{c}{\textbf{SEED-VIG}} \\
\midrule
\textbf{Methods} & \textbf{Year} & \textbf{ACC} & \textbf{Kappa} & \textbf{Methods} & \textbf{Year} & \textbf{ACC} & \textbf{Kappa} \\
\midrule
SVM & 2015 & 83.99 & 0.7912 & EEGNet & 2018 & 79.23 & 0.5928 \\
DGCNN & 2020 & 90.40 & 0.8560 & EEG-Conformer & 2023 & 89.66 & 0.6927 \\
V-IAG & 2021 & 95.64 & - & CSF-GTNet & 2024 & 81.48 & - \\
EEG-Conformer & 2023 & 95.30 & 0.9295 & SFT-Net & 2024 & 87.10 & 0.7170 \\
CU-GCN & 2024 & 95.70 & - & GAT-CNN & 2025 & 90.14 & - \\

\textbf{EEG-CSANet} & \textbf{2025} & \textbf{96.03} & \textbf{0.9404} & \textbf{EEG-CSANet} & \textbf{2025} & \textbf{90.56} & \textbf{0.7327} \\
\bottomrule
\end{tabular}
\end{table*}

The experimental results on BCIC-IV-2A are presented in Table~\ref{tab_3}. EEG-CSANet obtained an average accuracy of 88.54\%, representing an improvement of 15.10\% over the baseline model EEGNet and an increase of 3.51\% compared with the second-best method, EEG-TransNet. The average Kappa coefficient reached 0.8472, exceeding that of EEG-TransNet by 0.0468. Among the nine subjects, our model achieved the highest accuracy in seven cases, with a standard deviation of 8.41, indicating consistent performance across subjects. To statistically analyze the differences in decoding performance among all methods, we conducted paired-sample t-tests. Our approach showed a significant difference (p$<$0.05) compared with MSTFNet, and a very significant difference (p$<$0.01) compared with the other state-of-the-art methods.

The experimental results on BCIC-IV-2B are presented in Table~\ref{tab_4}. EEG-CSANet obtained an average accuracy of 91.09\%, representing an improvement of 10.28\% over the baseline model EEGNet and an increase of 1.39\% compared with the second-best method, TMSA-Net. The average Kappa coefficient reached 0.8218, exceeding that of TMSA-Net by 0.0278. Our model achieved the highest accuracy in seven subjects, while attaining the second-best accuracy in sub3 and sub6. Paired-sample t-test analysis further confirmed that the classification performance of our model was significantly different from that of other methods (p$<$0.05).

As shown in Table~\ref{tab_5}, on HGD, the EEG-CSANet outperformed several SOTA approaches, achieving an average accuracy of 97.15\%, which is 1.25\% higher than that of the second-best method, TMSA-Net. The average Kappa coefficient reached 0.9627, exceeding TMSA-Net by 0.018. In addition, the low STD of 2.97 indicates that the proposed model maintains stable and consistent classification performance across all 14 subjects.

To further validate the effectiveness of the proposed model, as shown in Table~\ref{tab_6}, we conducted additional experiments on two datasets for emotion recognition and fatigue monitoring (SEED and SEED-VIG).

On SEED, we compared EEG-CSANet with several representative methods evaluated under the same five-fold cross-validation protocol, including SVM\cite{zheng2015investigating}, DGCNN\cite{song2018eeg}, V-IAG\cite{song2021variational}, EEG-Conformer\cite{song2022eeg}, and CU-GCN\cite{gao2024graph}. EEG-CSANet achieved an average accuracy of 96.03\% and a Kappa coefficient of 94.04\%, outperforming all competing methods.

On SEED-VIG, comparisons were conducted with EEGNet\cite{lawhern2018eegnet}, CSF-GTNet\cite{gao2023csf}, SFT-Net\cite{gao2023sft}, EEG-Conformer\cite{song2022eeg}, and GAT-CNN\cite{chen2025driver}. EEG-CSANet obtained an average accuracy of 90.56\%, representing an improvement of 0.42\% over the second-best method, GAT-CNN. Notably, due to pronounced inter-subject variability, fatigue-related samples in SEED-VIG inevitably exhibit severe class imbalance. Under this challenging condition, EEG-CSANet still achieved a Kappa coefficient of 0.7327, exceeding SFT-Net by 0.0157, which demonstrates its robustness and reliability in imbalanced fatigue-detection scenarios.

In summary, without any additional tuning of model parameters, EEG-CSANet consistently achieves competitive performance across different experimental paradigms, demonstrating its strong transferability and robustness.

\section{Discussion}
\subsection{Ablation Experiments}
To investigate the contribution of different components in the proposed model, ablation experiments were conducted on three EEG datasets corresponding to distinct elicitation paradigms (BCIC-IV-2A, SEED, and SEED-VIG). As summarized in Table~\ref{tab_7}, we systematically evaluated the effects of the core modules, including S\&R, TCN, ResNet, and MSCA, as well as the Top-k and AvgPool components within the MSCA module, by removing or replacing each component in turn.

\begin{table}[t!]
\centering
\caption{Network Configurations for Ablation Experiments on Three EEG Datasets (BCIC-IV-2A, SEED, and SEED-VIG).}
\label{tab_7}
\small
\renewcommand{\arraystretch}{1.2}
\setlength{\tabcolsep}{7.5pt}
\begin{tabular}{ccccccc}
\toprule
\multirow{2}{*}{\textbf{Module}} & \multirow{2}{*}{\textbf{S\&R}} & \multirow{2}{*}{\textbf{TCN}} & \multirow{2}{*}{\textbf{Residual}} & \multicolumn{2}{c}{\textbf{MSCA}} \\
\cline{5-6}
 & & & & \textbf{Top-K} & \textbf{AvgPool} \\
\midrule
\textbf{Net1}    & \ding{51} & \ding{51} & \ding{51} & \ding{51} & \ding{51} \\
\textbf{Net2}    & \ding{55} & \ding{51} & \ding{51} & \ding{51} & \ding{51} \\
\textbf{Net3}    & \ding{51} & \ding{55} & \ding{51} & \ding{51} & \ding{51} \\
\textbf{Net4}    & \ding{51} & \ding{51} & \ding{55} & \ding{51} & \ding{51} \\
\textbf{Net5}    & \ding{51} & \ding{51} & \ding{51} & \ding{55} & \ding{55} \\
\textbf{Net6}    & \ding{51} & \ding{51} & \ding{51} & \ding{51} & \ding{55} \\
\textbf{Net7}    & \ding{51} & \ding{51} & \ding{51} & \ding{55} & \ding{51} \\
\bottomrule
\end{tabular}
\end{table}

\begin{figure*}[!t]
\centering
\includegraphics[width=7in]{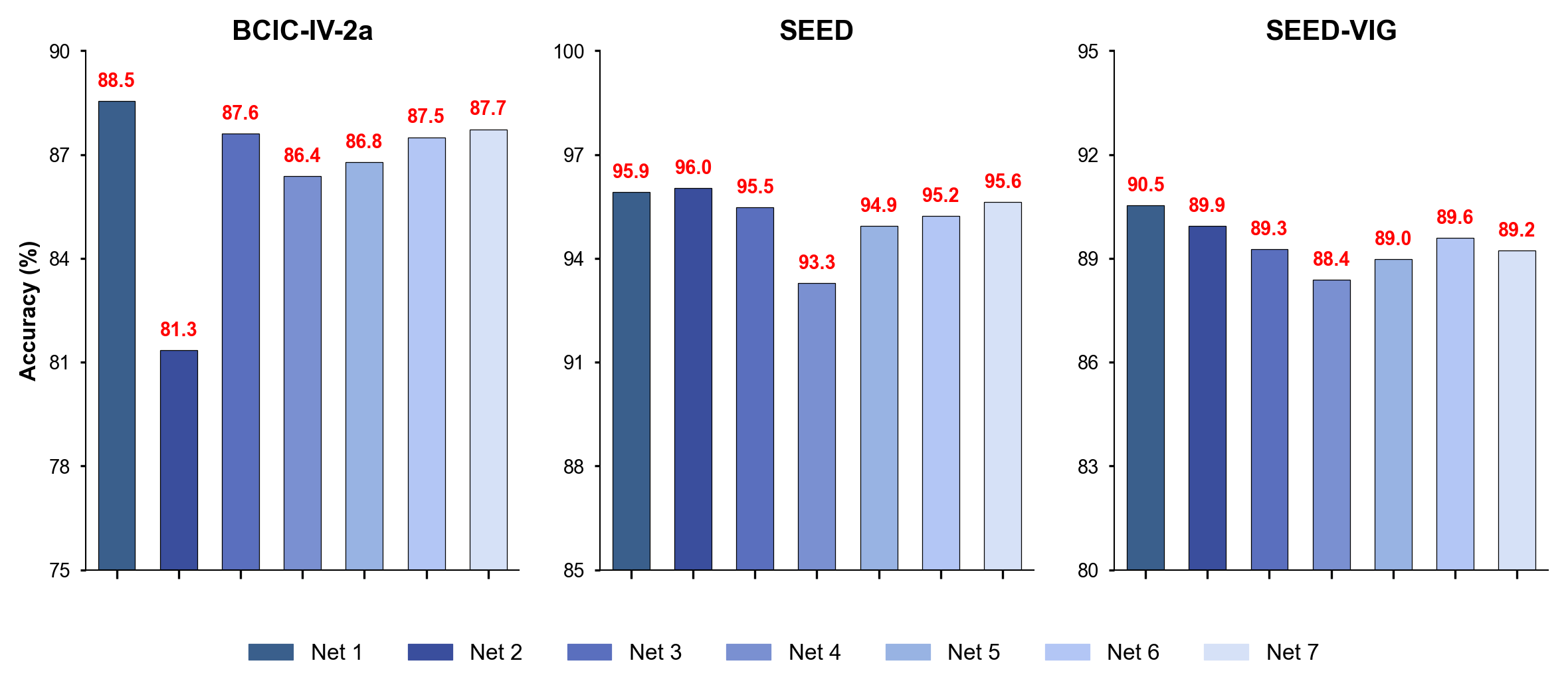}
\caption{Ablation experiments results on three EEG datasets (BCIC-IV-2A, SEED, and SEED-VIG).}
\label{fig_3}
\end{figure*}

As shown in Figure~\ref{fig_3}, the S\&R module improves performance by 7.19\% on BCIC-IV-2A, but only by 0.60\% on SEED-VIG, and slightly decreases accuracy ($-0.11\%$) on SEED. In addition, across all datasets, Net3, which excludes residual connections, shows a more noticeable drop in performance, suggesting that residual connections play an important role in maintaining model effectiveness. These findings indicate that the MSCA operation may be associated with partial loss of temporal feature information, whereas residual connections appear to help preserve and supplement such information, thereby supporting the representation of global features.

Motivated by the strong performance of the S\&R module on BCIC-IV-2A and its comparatively weaker results on SEED and SEED-VIG , we performed additional experiments on five datasets to further investigate the underlying performance differences.

\begin{table}[t!]
\centering
\caption{Performance Comparison of S\&R Method on Different EEG Datasets, Where W Denotes with Augmentation and N Denotes Without Augmentation.}
\label{tab_8}
\small
\renewcommand{\arraystretch}{1.2}
\setlength{\tabcolsep}{6.7pt}
\begin{tabular}{ccccccc}
\toprule
\multirow{2}{*}{\textbf{Dataset}} & \multicolumn{2}{c}{\textbf{ACC}} & \multicolumn{2}{c}{\textbf{STD}} & \multicolumn{2}{c}{\textbf{Kappa}} \\
\cline{2-7}
 & \textbf{W} & \textbf{N} & \textbf{W} & \textbf{N} & \textbf{W} & \textbf{N} \\
\midrule
BCIC-IV-2A     & 88.54 & 81.35 & 8.41 & 7.91 & 0.8472 & 0.7536 \\
BCIC-IV-2B     & 91.09 & 89.02 & 8.48 & 7.03 & 0.8218 & 0.7804 \\
HGD    & 97.15 & 93.75 & 2.97 & 5.37 & 0.9627 & 0.9167 \\
SEED   & 95.92 & 96.03 & 2.18 & 1.94 & 0.9401 & 0.9404 \\
SEED-VIG & 90.53 & 89.93 & 6.37 & 6.76 & 0.7327 & 0.7153 \\
\bottomrule
\end{tabular}
\end{table}

As shown in Table~\ref{tab_8}, the effectiveness of the S\&R module varies notably across datasets. On BCIC-IV-2A, BCIC-IV-2B, and HGD, S\&R yields stable improvements in both accuracy and Kappa, with the most significant gains observed on BCIC-IV-2A. This is likely due to the relatively small training sample sizes of these datasets, where complex models are more susceptible to overfitting, thereby yielding more pronounced performance gains. In SEED and SEED-VIG, although the overall model accuracy remains high, the introduction of S\&R results in only limited performance improvements, and some even exhibit a decline. These observations suggest that the practical benefits of this strategy in such tasks are rather limited.

Overall, the ablation results demonstrate that removing any module degrades performance, confirming that all components contribute complementarily to the effectiveness of the proposed architecture.

\subsection{Performance Across Different Branches}

\begin{figure}[!t]
\centering
\includegraphics[width=3.5in]{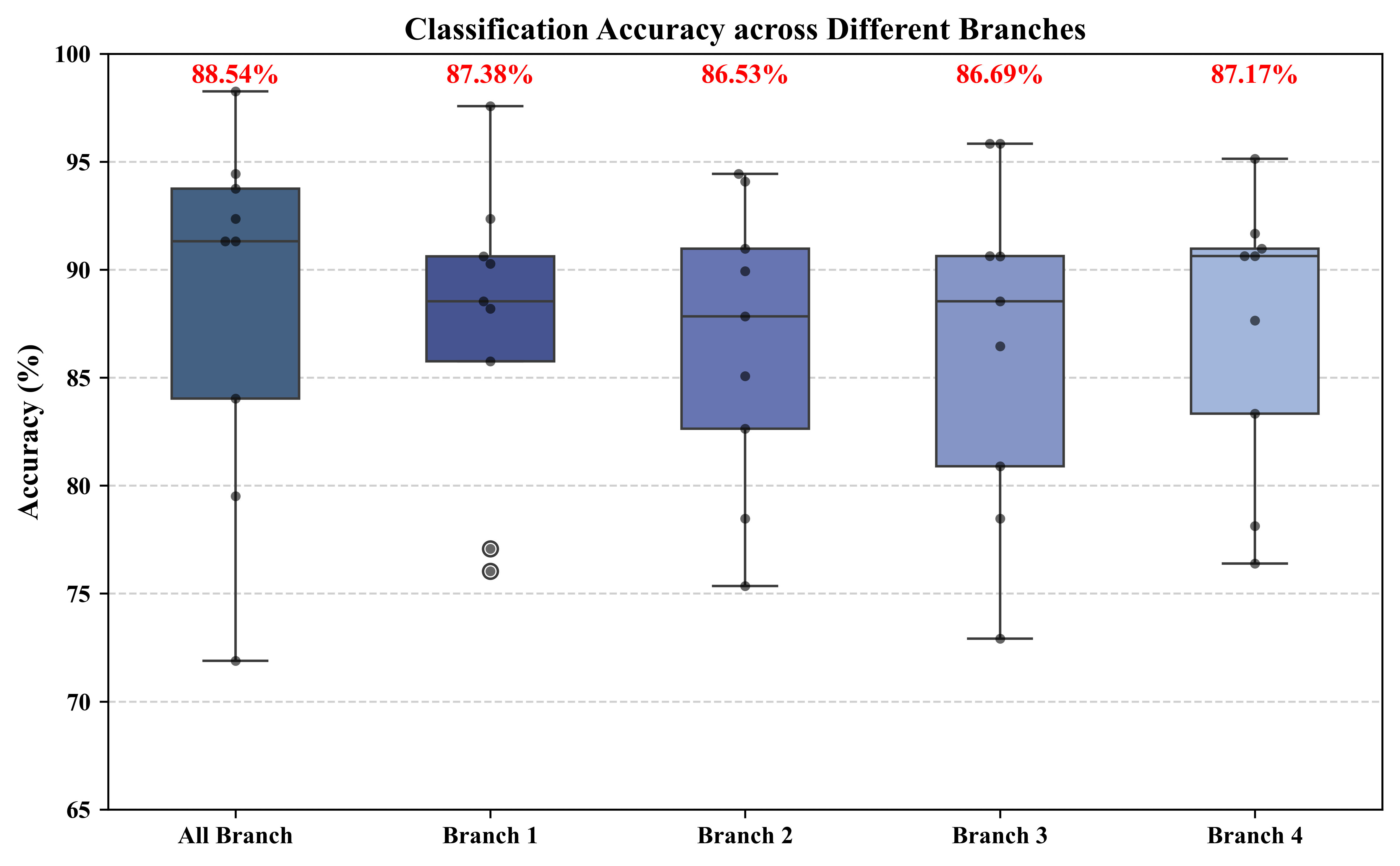}
\caption{Performance comparison of individual and fused branches in EEG-CSANet on BCIC-IV-2A}
\label{fig_4}
\end{figure}

To further analyze the effectiveness of the multi-branch structure in EEG-CSANet and the contribution of each branch to the overall performance, we conducted comparative experiments on BCIC-IV-2A. As shown in Figure~\ref{fig_4}, the results show that different branches contribute unequally to the model performance. Specifically, the first and fourth branches achieve higher average accuracy and provide more significant performance gains, whereas the second and third branches exhibit relatively lower individual performance and limited contributions. When all four branches are fused (All Branch), the classification performance becomes more concentrated in the high-accuracy range and shows a higher median value. This demonstrates that the multi-branch representations are complementary and that combining them improves the overall classification performance. Moreover, we found that even individual branches achieve relatively high performance on BCIC-IV-2A. Accordingly, we provide additional ablation experiments on the internal modules of each branch in the Appendix.

\subsection{Feature Visualization and Confusion Matrices}

\begin{figure*}[!t]
\centering
\includegraphics[width=6in]{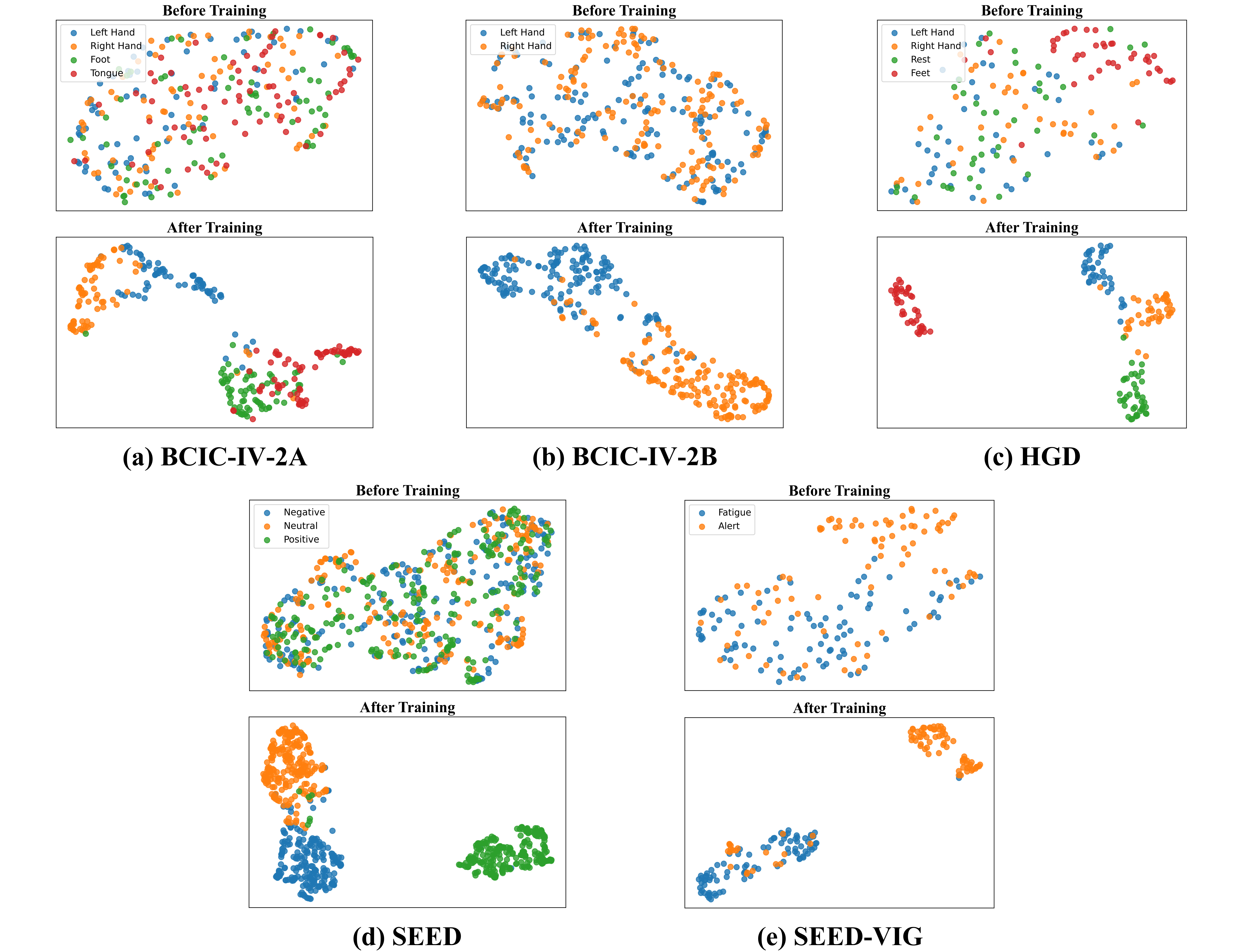}
\caption{The UMAP Visualization of EEG-CSANet Features Before and After Training. (a) BCIC-IV-2A for subject 1. (b) BCIC-IV-2B for subject 7. (c) HGD for subject 9. (d) SEED for subject 11. (e) SEED-VIG for subject 18.}
\label{fig_5}
\end{figure*}

In order to further illustrate the efficacy of EEG-CSANet, the Uniform Manifold Approximation and Projection (UMAP) method is applied to visualize the classification features extracted from five datasets. Compared with t-SNE, UMAP offers higher computational efficiency and scalability for high-dimensional data, revealing clustering patterns and distribution characteristics in low-dimensional space\cite{mcinnes2018umap}. As illustrated in Figure~\ref{fig_5}, prior to training, the extracted features are dispersed in the low-dimensional space, exhibiting substantial overlap between different classes and limited overall discriminative capability. Following model training, the feature distribution forms a more compact and orderly clustering structure, with markedly enhanced inter-class separability.

\begin{figure*}[!t]
\centering
\includegraphics[width=7in]{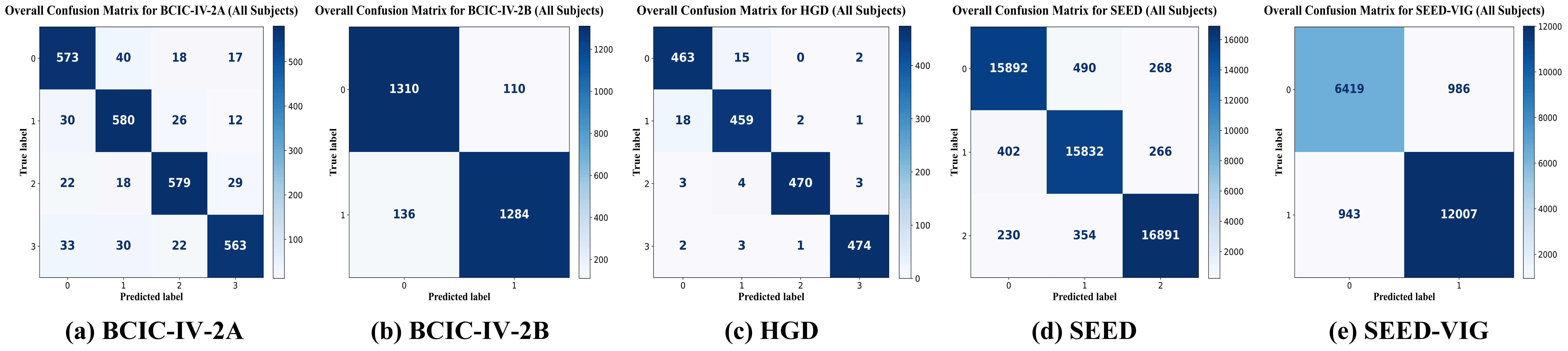}
\caption{Confusion matrix of EEG-CSANet on the five datasets. (a) BCIC-IV-2A. (b) BCIC-IV-2B. (c) HGD. (d) SEED. (e) SEED-VIG.}
\label{fig_6}
\end{figure*}

Moreover, Figure~\ref{fig_6} shows the confusion matrix for each class with EEG-CSANet on five datasets. The confusion matrix is a fundamental tool for evaluating classification models, providing insights into accuracy, error patterns, and class-specific biases for a comprehensive performance assessment. 

In Figure~\ref{fig_6}.a (BCIC-IV-2A), the classification performance for tongue movement (task 3) is slightly lower compared to other classes. In Figure~\ref{fig_6}.c (HGD), the classification accuracy for left-hand (task 0) and right-hand (task 1) movements is relatively weaker. In Figure~\ref{fig_6}.e (SEED-VIG), due to the substantially smaller number of awake samples compared to fatigued samples, the recognition performance for the awake state (task 0) is somewhat affected. Nevertheless, the model demonstrates overall balanced performance across different classes in all datasets, without exhibiting a pronounced bias toward any specific class.

\section{Conclusion}
In this paper, we propose EEG-CSANet, a multi-branch feature fusion framework for EEG decoding. To address the spatial information loss caused by shared weights in multiscale feature extraction, EEG-CSANet introduces a parallel design that assigns an independent spatial module to each temporal scale, enabling more precise characterization of the spatiotemporal heterogeneity of EEG signals. For multi-branch feature fusion, EEG-CSANet adopts a main–auxiliary collaborative architecture: the main branch leverages multiscale self-attention to model core spatiotemporal patterns, while the auxiliary branch employs multiscale sparse cross-attention to enable efficient local interactions with the main branch. Experimental results demonstrate that EEG-CSANet achieves SOTA performance across five public EEG datasets. Moving forward, we expect EEG decoding to to be more closely aligned with neural mechanism-driven strategies, fostering the design of methods specifically adapted to the properties of EEG signals, which in turn can improve the physiological plausibility and generalizability of EEG-based models.

\section*{REFERENCES}
\vspace{-1.5em}
\bibliography{literature}
\bibliographystyle{IEEEtran}
\vfill

\clearpage
\onecolumn
\begin{center}
\Large\bfseries Appendix
\end{center}

\addcontentsline{toc}{section}{Appendix}
\subsection{Baselines}
\textbf{EEGNet}~\cite{lawhern2018eegnet} is a lightweight convolutional neural network specifically tailored for EEG signal classification. By leveraging depthwise separable convolutions, it achieves an effective trade-off between performance and computational efficiency. Its architecture is simple, compact, and easy to implement or transfer across different tasks. Owing to these advantages, EEGNet has been widely adopted in various BCI applications, such as MI and Event-Related Potention decoding, and has become one of the benchmark models in deep learning for EEG analysis.

\textbf{EEG-Conformer}~\cite{song2022eeg} is a deep learning model specifically developed for EEG signal classification, which integrates CNNs with a Transformer architecture. In this framework, CNNs are employed to extract local spatiotemporal features, while the Transformer is utilized to model long-range temporal dependencies and global contextual information. By effectively combining the strengths of feature extraction and sequence modeling, EEG-Conformer provides a powerful and comprehensive representation of EEG data.

\textbf{ATCNet}~\cite{altaheri2022physics} is a compact and interpretable attention-based temporal convolutional network specifically designed for EEG motor imagery classification. It integrates principles of scientific machine learning by employing convolutional layers to extract spatiotemporal features, while multi-head self-attention mechanisms emphasize the most informative temporal segments. In addition, a TCN is incorporated to capture long-term temporal dependencies. With its concise architecture and strong interpretability, ATCNet demonstrates improved decoding performance for EEG signals.

\textbf{ADFCNN}~\cite{tao2023adfcnn} is a deep learning model tailored for BCI. It adopts a dual-scale convolutional architecture to separately extract temporal features of EEG rhythms as well as global and fine-grained spatial features. To further enhance representation learning, a self-attention mechanism is introduced, enabling the model to dynamically weight and fuse multiscale information according to intrinsic feature similarities, thereby improving feature discriminability. This design effectively overcomes the limitations of traditional single-scale or traditional multiscale CNNs in feature extraction and fusion, offering stronger modeling capabilities for both spectral and spatial information.

\textbf{EEG-Transnet}~\cite{ma2024attention} is a EEG classification model that integrates CNNs with a self-attention mechanism. The model extracts multimodal temporal features from both the mean and variance dimensions, while a shared self-attention module captures global temporal dependencies. A convolutional encoder is then employed to fuse these features, thereby enhancing their discriminability. In addition, a signal S\&R-based data augmentation strategy is introduced to further improve robustness and decoding performance.

\textbf{MSTFNet}~\cite{jin2024multiscale} is an end-to-end convolutional neural network proposed for motor imagery EEG classification. To overcome the limitations of traditional single-scale convolutions in feature extraction, MSTFNet is designed with four core modules: feature enhancement, multiscale temporal feature extraction, spatial feature extraction, and feature fusion. Through the collaborative extraction of multiscale spatiotemporal features and the adoption of refined fusion strategies, the model substantially improves the decoding of EEG signals.

\textbf{EISATC-Fusion}~\cite{liang2024eisatc} is a neural network model tailored to EEG decoding, which integrates Inception modules, multi-head self-attention, TCNs, and a layer fusion structure. Specifically, the model employs DS-Inception to extract multiscale frequency band features, incorporates cnnCosMSA to alleviate attention collapse and enhance interpretability, and applies depthwise separable convolutions to reduce parameter complexity. Furthermore, by combining both feature-level and decision-level fusion strategies, EISATC-Fusion achieves improved robustness in EEG signal decoding.

\textbf{MCMTNet}~\cite{yang2025mcmtnet} is a deep learning model capable of directly processing raw EEG signals without the need for complex preprocessing. Its architecture is composed of three main components: a multi-domain convolution module, a multi-head attention module, and a TCN. The multi-domain convolution module is responsible for filtering and feature extraction, the multi-head attention module enhances cross-scale correlations, and the TCN improves temporal coherence.

\textbf{TMSA-Net}~\cite{zhao2025tmsa} is a neural network model that integrates CNNs with an improved Transformer-based attention mechanism. The model first employs CNNs to extract local spatiotemporal features, and then introduces a novel attention module to enhance global modeling capability across both the channel and temporal dimensions, thereby effectively bridging the gap between local and global representations. By optimizing the attention structure, TMSA-Net not only reduces computational overhead but also improves feature fusion efficiency, enhancing the model's sensitivity to key EEG patterns and its interpretability.

\textbf{SVM}\cite{zheng2015investigating} is a classical supervised learning model widely employed in small-sample, high-dimensional pattern recognition tasks such as EEG signal classification. Its core principle lies in finding an optimal hyperplane that maximizes the margin between samples of different classes, thereby yielding a decision boundary with strong generalization capability. In EEG analysis, SVM is typically combined with specific feature extraction methods such as time-frequency features or Common Spatial Patterns (CSP) and has demonstrated robust performance in EEG decoding tasks.

\textbf{DGCNN}\cite{song2018eeg} is a dynamic graph convolutional neural network specifically designed for modeling EEG signals. Unlike traditional graph convolutional networks that rely on predefined or fixed adjacency matrices, DGCNN dynamically learns the connection weights between EEG channels during training, thereby adaptively constructing an adjacency matrix that captures the intrinsic functional relationships among channels. Thanks to its flexible graph structure learning capability and end-to-end trainability, DGCNN has demonstrated superior performance in various EEG analysis tasks such as emotion recognition and cognitive state decoding, and has become one of the key benchmark approaches for processing EEG signals with graph neural networks.

\textbf{V-IAG}\cite{song2021variational} is a graph-based modeling approach designed for EEG emotion recognition, aiming to jointly capture individual differences and dynamic uncertainties among brain regions. The method comprises two branches: an instance-adaptive branch that dynamically constructs a graph reflecting subject-specific channel dependencies based on the input, and a variational branch that generates a probabilistic graph to quantify connection uncertainty. The fused graph is then used to extract more discriminative features. To further enhance representation capability, V-IAG introduces a multi-level multi-graph convolution operation that aggregates channel features across different frequency bands using distinct graph structures, complemented by graph coarsening and sparsity constraints to improve robustness.

\textbf{CU-GCN}\cite{gao2024graph} is a graph convolutional network for EEG emotion recognition that employs distribution-based uncertainty modeling to adaptively weight node features, effectively alleviating over-smoothing in long-range propagation while capturing spatial dependencies and temporal-spectral relationships among EEG channels. The model integrates graph mixup to enhance latent connections and suppress label noise, and incorporates the uncertainty mechanism into deep GCN weights via a one-way learning strategy, significantly outperforming all baseline models.

\textbf{CSF-GTNet}\cite{gao2023csf} is a multi-dimensional feature fusion network designed for driver fatigue detection, which innovatively integrates time-domain and spatial-spectral information to enhance the representation capability of EEG signals. The model consists of two parallel branches: the Gaussian Time-domain Network (GTNet), which adaptively captures fatigue-related temporal dynamics, and the pure Convolutional Spatial-Frequency Network (CSFNet), which focuses on extracting multi-scale spatial distributions and spectral characteristics. By fusing features from these complementary dimensions, CSF-GTNet provides a more comprehensive characterization of the differences in EEG patterns between alert and fatigue states.

\textbf{SFT-Net}\cite{gao2023sft} is a lightweight and efficient model for EEG-based driver fatigue detection that enhances recognition performance by fusing spatial, spectral, and temporal information. The method constructs a 4D feature tensor from the differential entropy of five frequency bands, employs an attention mechanism to recalibrate spatial and spectral features for each time frame, extracts key representations using depthwise separable convolution, and captures temporal dependencies via LSTM. SFT-Net outperforms existing methods and demonstrates strong interpretability, validating the effectiveness of multi-dimensional information fusion in fatigue detection.

\textbf{GAT-CNN}\cite{chen2025driver} is an end-to-end EEG model for driver fatigue detection that enhances recognition performance by explicitly modeling dependencies among EEG channels. The method takes preprocessed EEG signals and an adjacency matrix constructed from mutual information as input, uses a graph attention mechanism to adaptively learn inter-channel weights, and combines convolutional operations to extract structured spatiotemporal features without manual feature engineering. Alert and fatigue states are finally classified using fully connected layers and a Softmax classifier. GAT-CNN demonstrates excellent predictive accuracy and generalization capability in experiments.

\begin{table*}[!t]
\centering
\caption{Performance Comparison on HGD.}
\label{tab_9}
\small
{\fontsize{7}{8.2}\selectfont 
\resizebox{\textwidth}{!}{
\begin{tabular}{ccccccccc}
\toprule
\textbf{Subjects} & \textbf{EEGNet} & \textbf{EEG-Conformer} & \textbf{ATCNet} & \textbf{EEG-TransNet} & \textbf{ADFCNN} & \textbf{MCMTNet} & \textbf{TMSA-Net} & \textbf{EEG-CSANet} \\
\midrule
H01 & 85.15 & 92.28 & 95.00 & 94.38 & 90.00 & 95.69 & \textbf{98.13} & 97.50 \\
H02 & 85.02 & 89.78 & \textbf{97.50} & 95.00 & 92.50 & 97.00 & 94.38 & \textbf{97.50} \\
H03 & 98.82 & 98.23 & 99.38 & 99.38 & 99.17 & 99.75 & 99.38 &  \textbf{100.00} \\
H04 & 95.70 & 98.85 & 98.75 & 99.38 & \textbf{100.00} & 98.31 & \textbf{100.00} & 99.38 \\
H05 & 93.20 & 90.73 & 97.50 & 94.38 & \textbf{100.00} & 97.94 & \textbf{100.00} & 99.38 \\
H06 & 90.12 & 93.33 & 95.00 & 95.00 & 98.13 & 97.81 & \textbf{98.75} & \textbf{98.75} \\
H07 & 86.28 & 90.04 & 94.38 & 94.97 & 92.50 & 93.77 & \textbf{98.75} & 95.63 \\
H08 & 88.82 & 85.10 & 96.88 & 94.38 & \textbf{99.38} & 95.87 & 96.88 & 96.88 \\
H09 & 95.07 & 98.23 & 98.13 & 97.50 & \textbf{100.00} & 97.62 & 99.38 & 99.38 \\
H10 & 88.22 & 90.73 & 91.88 & 95.63 & \textbf{96.25} & 93.81 & 95.00 & \textbf{96.25} \\
H11 & 76.35 & 79.58 & 85.63 & 83.13 & \textbf{98.13} & 88.88 & 98.75 & 92.63 \\
H12 & 96.95 & 96.45 & 98.13 & 98.75 & 98.13 & 98.06 & 98.75 & \textbf{98.75} \\
H13 & 83.75 & 92.65 & 95.00 & 95.60 & 97.50 & 96.10 & 97.50 & \textbf{98.75} \\
H14 & 80.72 & 82.05 & \textbf{94.38} & 90.00 & 70.63 & 89.56 & 66.88 & 89.38 \\
\midrule
\textbf{Acc} & 88.87 & 91.29 & 95.54 & 94.82 & 95.17 & 95.73 & 95.90 & \textbf{97.15} \\
\textbf{Std}     &  6.53 &  5.90 &  3.54 &  4.17 &  7.74 &  3.23 &  8.52 & \textbf{2.97} \\
\textbf{Kappa}   & 0.8531 & 0.8824 & 0.9405 & 0.9309 & 0.9381 & 0.9430 & 0.9452 & \textbf{0.9627} \\
\bottomrule
\end{tabular}}}
\end{table*}

\subsection{Detailed Experiment Results on HGD}
Table~\ref{tab_9} reports the subject-dependent performance comparisons on HGD. Specifically, we present a comprehensive comparison of classification accuracy across all individual subjects between our method and several state-of-the-art (SOTA) models, highlighting the consistency and competitiveness of our approach at the subject level.

\subsection{Subject-independent Experiments on BCIC-IV-2A}
To further explore the generalization potential of the proposed model, we conducted an additional subject-independent classification experiment on BCIC-IV-2A. In our subject-independent experiment, we adopted the leave-one-subject-out strategy, using the complete data from eight subjects for training while reserving all data from the ninth subject exclusively for testing. Table~\ref{tab_10} presents a performance comparison between EEG-CSANet and several SOTA methods under this setting. Although this experiment was not the primary focus of our study, the results show that EEG-CSANet still achieves more competitive classification accuracy, providing preliminary validation of its transferability in cross-subject scenarios. We plan to systematically enhance the model's adaptability in cross-subject and cross-session settings and investigate its feasibility for deployment in real-world brain-computer interface applications.

\begin{table*}[!t]
\centering
\caption{Performance comparison on BCIC-IV-2a (subject-independent).}
\label{tab_10}
\large
{\fontsize{8}{10}\selectfont 
\setlength{\tabcolsep}{10pt}
\begin{tabular}{ccccc}
\toprule
\textbf{Methods} & \textbf{Year} & \textbf{Acc} & \textbf{Kappa} \\
\midrule
EEGNet           & 2018 & 58.19 & 0.4425 \\
ATCNet           & 2022 & 62.85 & 0.5047 \\
EEG-TransNet     & 2024 & 65.73 & 0.5430 \\
MCMTNet          & 2025 & 66.31 & 0.5508 \\
\midrule
\textbf{EEG-CSANet} & \textbf{2025} & \textbf{69.68} & \textbf{0.5957} \\
\bottomrule
\end{tabular}}
\end{table*}

\subsection{Ablation Experiments within Single Branch}
\begin{figure}[!t]
\centering
\includegraphics[width=3.5in]{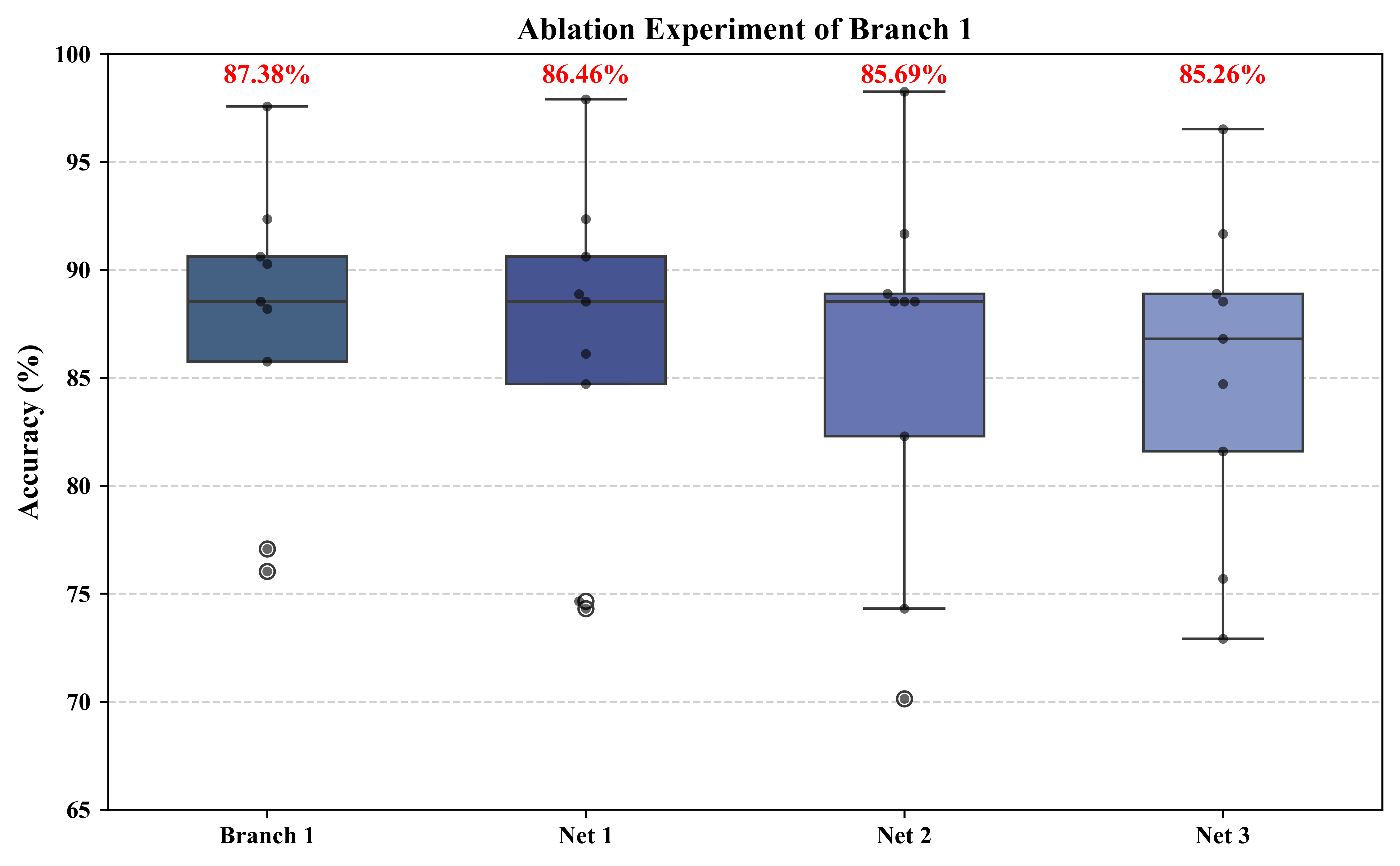}
\caption{Ablation experiments of Branch 1 on BCIC-IV-2A. Net 1: Moving AvgPool. Net 2: Moving Residual.  Net 3: Moving AvgPool and Residual.}
\label{fig_7}
\end{figure}

As shown in Figure~\ref{fig_4}, even under a single-branch configuration, EEG-CSANet achieves competitive classification accuracy on BCIC-IV-2A, with the first branch demonstrating particularly strong performance. To investigate this further, we isolate the first branch and conduct ablation experiments within the single branch, as illustrated in Figure~\ref{fig_7}. This study focuses on two core components: the AvgPool operation in MSA and the Residual connection structure, both of which represent key innovations in our model. The results indicate that both AvgPool and the residual network significantly impact model performance, with the residual connection showing a notably stronger contribution. Therefore, we argue that the residual connection not only effectively mitigates feature degradation during deep propagation but also enhances feature reuse, making it a critical factor in improving the performance of both the single branch and the overall model.

\subsection{Hierarchical Guidance of Structure}
\begin{figure}[!t]
    \centering
    \includegraphics[width=5in]{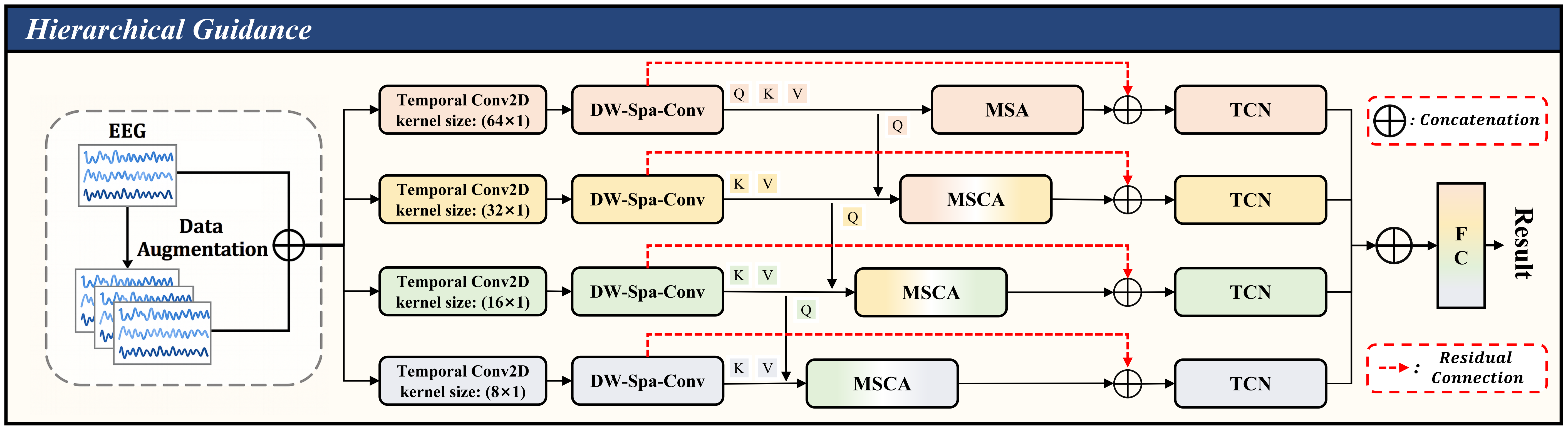}
    \caption{Hierarchical Guidance of Structure.}
    \label{fig_08}
\end{figure}

Inspired by~\cite{liu2025vieeg}, we compared the performance of the main--auxiliary structure from EEG-CSANet with a hierarchical structure, as is shown in Figure~\ref{fig_08}. In the hierarchical design, the features extracted after the DW-Spa-Conv module in the first branch serve as the Query for the second branch's MSCA, whose output is then passed sequentially as the Query to the third and fourth branches. This layer-by-layer propagation enables each branch to attend not only to its own representations but also to inherit information from the preceding branch, thereby achieving cross-branch multi-level feature fusion. As is shown in Table~\ref{tab_11}, compared with main--auxiliary structure, the Hierarchical structure yields slightly lower classification performance, though the difference is not statistically significant according to t-tests. To present the best-performing model, we ultimately adopted the main--auxiliary structure.

\label{app:cengji}

\begin{table}[!t]
\centering
\caption{Comparison between Main-auxiliary Guidance and Hierarchical Guidance.}
\label{tab_11}
{\fontsize{8}{10}\selectfont 
\setlength{\tabcolsep}{10pt}
\begin{tabular}{ccccccc}
\toprule
\multirow{2}{*}[-0.7ex]{\textbf{Dataset}} & \multicolumn{3}{c}{\textbf{Main–auxiliary Guidance}} & \multicolumn{3}{c}{\textbf{Hierarchical Guidance}} \\
\cmidrule(lr){2-4} \cmidrule(lr){5-7}
 & \textbf{ACC} & \textbf{Std} & \textbf{Kappa} & \textbf{ACC} & \textbf{Std} & \textbf{Kappa} \\
\midrule
BCIC-IV-2A & 88.54 & 8.31 & 0.8472 & 88.08 & 8.77 & 0.8411 \\
BCIC-IV-2B & 91.09 & 8.48 & 0.8218 & 90.89 & 8.57 & 0.8178 \\
HGD        & 97.15 & 2.97 & 0.9627 & 96.29 & 4.01 & 0.9503 \\
\bottomrule
\end{tabular}}
\end{table}

\begin{figure}[!t]
    \centering
    \includegraphics[width=4in]{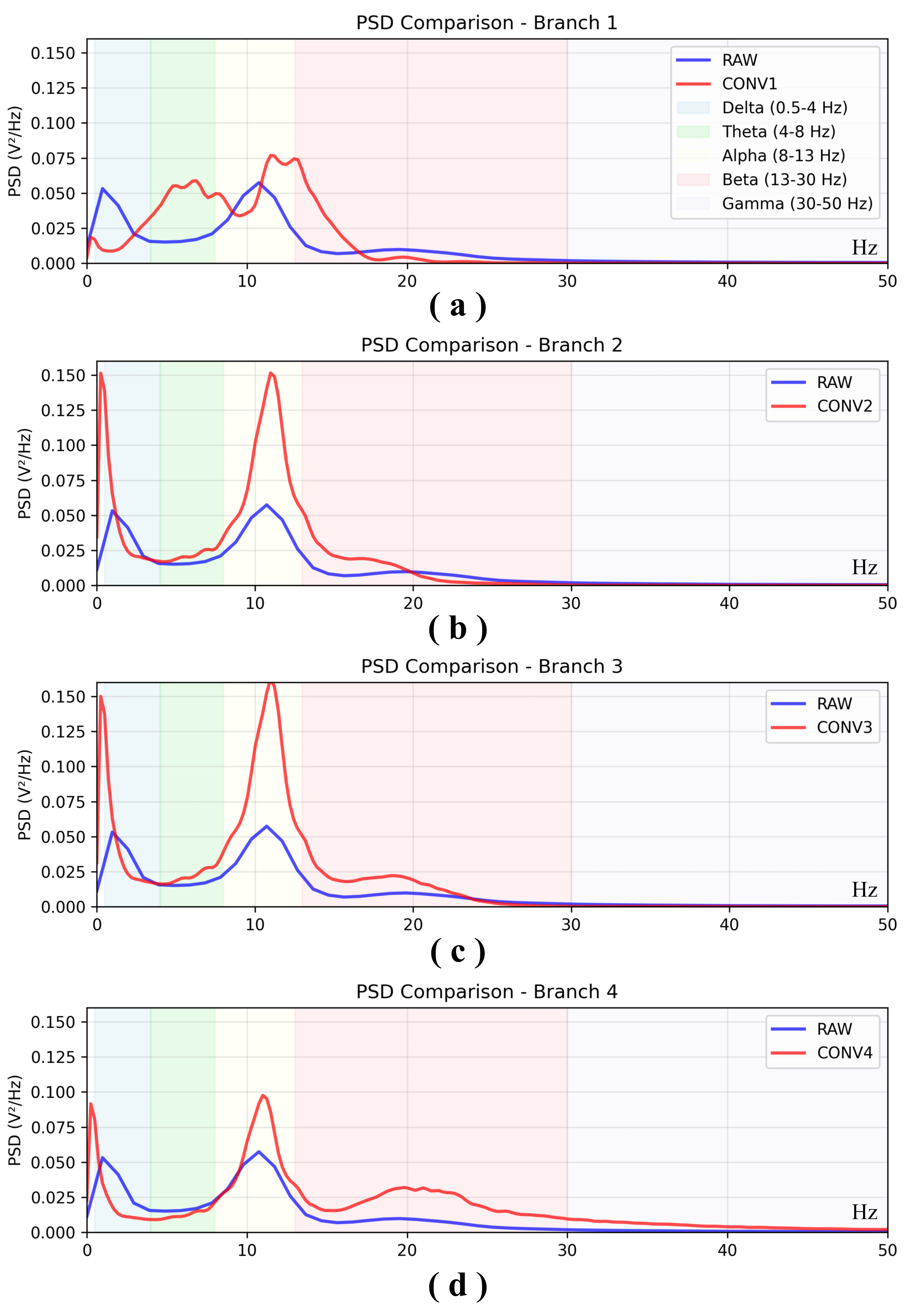}
    \caption{PSD of EEG-CSANet on four branches (HGD: Sub6).}
    \label{fig_9}
\end{figure}

\subsection{Visualization of Convolutional Features}
Finally, to further investigate the features extracted by the four branches using convolutional kernels of different sizes, we applied the welch method to compute the power spectral density (PSD) of the EEG signals after temporal convolution. From top to bottom, convolutional kernel sizes of 64, 32, 16, and 8 were used. In Figure~\ref{fig_9}, the blue line represents the original signal, while the red line represents the convolved signal. Figure~\ref{fig_9}.a shows a pronounced energy enhancement in the Theta, Alpha, and Beta bands, particularly in the 8--13 Hz and 13--20 Hz ranges, whereas the Delta band exhibits a marked decrease in energy. Figure~\ref{fig_9}.b exhibits a sharp peak in the Alpha band, with a notable increase in the Delta band, but minimal enhancement in the high-frequency range. Figure~\ref{fig_9}.c is similar to Figure~\ref{fig_9}.b but shows better representation in the transition from Alpha to Beta. Figure~\ref{fig_9}.d displays relatively weaker low-frequency activity (0.5--13 Hz) compared to Figure~\ref{fig_9}.b and Figure~\ref{fig_9}.c, but shows marked enhancement in the high-frequency Beta to Gamma (13--50 Hz) range, with a peak around 20 Hz. 

This may be attributed to the larger receptive field of the large kernels, which can capture the overall slow-varying trends of the signal and cross-frequency rhythmic information due to their coverage of longer temporal windows. Conversely, smaller kernels focus on local, rapidly varying signal patterns, making them more sensitive to high-frequency features of EEG signals, which often reflect transient, short-duration neural activity. 

Therefore, the multi-branch temporal convolutional module indeed captures features from different frequency bands, and features in different frequency bands often correspond to distinct spatial distribution patterns. This further validates the advantage of our multi-branch architecture: by separately extracting multi-scale features along the temporal dimension while preserving their independent spatial representation pathways, it effectively mitigates the issue of spatial information confusion or loss that arises in traditional approaches when multi-scale features are directly concatenated.

\end{document}